\documentclass{ecai}

\usepackage{times}
\usepackage{graphicx}
\usepackage{latexsym}

\usepackage[utf8]{inputenc}

\usepackage{xcolor}
\usepackage{colortbl}
\definecolor{LightCyan}{rgb}{0.88,1,1}
\definecolor{LightGray}{rgb}{0.9,0.9,0.9}

\usepackage{dsfont}
\usepackage{amsmath}

\usepackage[noend]{algpseudocode}
\usepackage{amssymb}
\usepackage{amsthm}
\usepackage{bbm}

\usepackage{hyperref}
\algnewcommand\And{\textbf{and}} 

\usepackage{caption}
\usepackage{subcaption}

\begin{document}

\title{MDE: Multiple Distance Embeddings for Link Prediction in Knowledge Graphs}
\author{%
    Afshin Sadeghi\institute{Smart Data Analytics Group, University of Bonn, Germany; Fraunhofer IAIS, Germany, email: sadeghi@cs.uni-bonn.de}
    \and
    Damien Graux\institute{ADAPT Centre, Trinity College Dublin, Ireland, email: grauxd@tcd.ie}
    \and
    Hamed Shariat Yazdi\institute{Smart Data Analytics Group, University of Bonn, Germany, email: shariat@cs.uni-bonn.de}
    \and
    Jens Lehmann\institute{Smart Data Analytics Group, University of Bonn, Germany; Fraunhofer IAIS, Germany, email: jens.lehmann@iais.fraunhofer.de}
}
\maketitle

\begin{abstract}
Over the past decade, knowledge graphs became popular for capturing structured domain knowledge. 
Relational learning models enable the prediction of missing links inside knowledge graphs. More specifically, latent distance approaches model the relationships among entities via a distance between latent representations.
Translating embedding models (e.g., TransE) are among the most popular latent distance approaches which use one distance function to learn multiple relation patterns. 
However, they are mostly inefficient in capturing symmetric relations since the representation vector norm for all the symmetric relations becomes equal to zero. They also lose information when learning relations with reflexive patterns since they become symmetric and transitive.
We propose the Multiple Distance Embedding model (MDE) that addresses these limitations and a framework to collaboratively combine variant latent distance-based terms.
Our solution is based on two principles: 1) we use a limit-based loss instead of a margin ranking loss and, 2) by learning independent embedding vectors for each of the terms we can collectively train and predict using contradicting distance terms.
We further demonstrate that MDE allows modeling relations with (anti)symmetry, inversion, and composition patterns. We propose MDE as a neural network model that allows us to map non-linear relations between the embedding vectors and the expected output of the score function.
Our empirical results show that MDE performs competitively to state-of-the-art embedding models on several benchmark datasets.

\end{abstract}

\section{Introduction}
While machine learning methods conventionally model functions given sample inputs and outputs, a subset of Statistical Relational Learning (SRL)~\cite{de2008logical,nickel2015review} approaches specifically aim to model ``things'' (entities) and relations between them. These methods usually model human knowledge which is structured in the form of multi-relational Knowledge Graphs (KG). KGs allow semantically rich queries and are used in search engines, natural language processing (NLP) and dialog systems. However, they usually miss many of the true relations~\cite{west2014knowledge}, therefore, the prediction of missing links/relations in KGs is a crucial challenge for SRL approaches.  

Practically, a KG usually consists of a set of facts. And a fact is a triple (head, relation, tail) where heads and tails are called entities.
Among the SRL models, distance-based KG embeddings are popular because of their simplicity, their low number of parameters, and their efficiency on large scale datasets. Specifically, their simplicity allows integrating them into many models. Previous studies have integrated them with logical rule embeddings~\cite{guo2016jointly}, have adopted them to encode temporal information~\cite{jiang2016encoding} and have applied them to find equivalent entities between multi-language datasets~\cite{kgmerging}. 

Soon after the introduction of the first multi-relational distance-based method TransE~\cite{bordes2013translating}, it was acknowledged that it is inefficient in learning symmetric relations, since the norm of the representation vector for all the symmetric relations in the KG becomes close to zero. This means the model cannot distinguish well different symmetric relations in a KG. 
To extend this model many variations were studied afterwards, e.g., TransH~\cite{wang2014knowledgeTransH}, 
TransR~\cite{lin2015learningTransR}, TransD~\cite{ji2015knowledgeTransD}, and STransE~\cite{StransE}. 
Even though they solved the issue of symmetric relations, they introduced an other limitation: these models were no longer efficient in learning the inversion and composition relation patterns that originally TransE could handle. 
 
Besides, as noted in~\cite{kazemi2018simple,sun2019rotate}, within the family of distance-based embeddings, reflexive relations are usually forced to become symmetric and transitive. In this study, we take advantage of independent vector representations of vectors that enable us to view the same relations from different aspects and put forward a translation-based model that addresses these limitations and allows the learning of all three relation patterns. 
In addition, we address the issue of the limit-based loss function in finding an optimal limit, and suggest an updating limit loss function to be used complementarily to the current limit-based loss function which has fixed limits. 
Moreover, we frame our model into a neural network structure that allows it to learn non-linear patterns for the limits in the limit based loss, improving the generalization power of the model in link prediction tasks.

The model performs well in the empirical evaluations, competing against state-of-the-art models in link prediction benchmarks. In particular, it outperforms\footnote{The complete code and the experimental datasets are available from: \url{https://github.com/mlwin-de/MDE}} state-of-the-art models on Countries~\cite{bouchard2015approximate} benchmark which is designed to evaluate composition pattern inference and modeling.

Since our approach involves several elements that model the relations between entities as the geometric distance of vectors from different views, we dubbed it \textbf{m}ultiple-\textbf{d}istance \textbf{e}mbeddings (MDE). 

The rest of this article is structured as follows: we define background and notations in Section~\ref{sec:Background and Notation} and summarize related efforts in Section~\ref{sec:Related Work}. Then we present the MDE model in Section~\ref{sec:MDE Multiple Distance Embeddings} and describes the extensions of the model including a hyperparameter search algorithm for the loss function and a Neural Network framing of MDE in Section~\ref{sec:extensions}. We report on the experiments in Section~\ref{sec:experiments} before concluding.




\section{Background and Notation}
\label{sec:Background and Notation}
Given the set of all entities $\mathcal{E}$ and the set of all relations $\mathcal{R}$, we formally define a fact as a triple of the form $(\mathbf{h}, \mathbf{r}, \mathbf{t})$ in which $\mathbf{\mathbf{h}}$ is the head and $\mathbf{t}$ is the tail, $\mathbf{h,t} \in \mathcal{E}$ and $\mathbf{r} \in \mathcal{R}$ is a relation. A knowledge graph $\mathcal{KG}$ is a subset of all true facts $\mathcal{KG} \subset \zeta$ and is represented by a set of triples.
An embedding is a mapping from an entity or a relation to their latent representation. A latent representation is usually a (set of) vector(s), a matrix or a tensor of numbers. A relational learning model is made of an embedding function and a prediction function that given a triple $(\mathbf{h}, \mathbf{r}, \mathbf{t})$ it determines if $(\mathbf{h}, \mathbf{r}, \mathbf{t}) \in \zeta$. We represent the embedding representation of an entity $\mathbf{h}$ with a lowercase letter $h$ if it is a vector and with an uppercase letter $H$ if it is a matrix.  
The ability to encode different patterns in the relations can show the generalization power of a model:

\textbf{Definition 1}. A relation $r$ is symmetric (antisymmetric) if $\forall x, y$
\begin{align*}
\  r(x, y) \Rightarrow r(y, x) \;\;(\  r(x, y) \Rightarrow \neg r(y, x) \ ).
\end{align*}

\textbf{Definition 2}. A relation $r_1$ is inverse to relation $r_2$ if $\forall x, y$
\begin{align*}
r_2(x, y) \Rightarrow  r_1(y, x). 
\end{align*}

\textbf{Definition 3}. A relation $r_1$ is composed of relation $r_2$ and 
relation $r_3$ if $\forall x, y, z $
\begin{align*}
r_2(x, y) \wedge r_3(y, z)  \Rightarrow r_1(x, z)
\end{align*}

\section{Related Work}
\label{sec:Related Work}

\textbf{Tensor Factorization and Multiplicative Models} define the score of triples via pairwise multiplication of embeddings. DistMult~\cite{dismult} simply multiplies the embedding vectors of a triple element by element $\langle h,r,t \rangle$ as the score function. Since multiplication of real numbers is symmetric, DistMult can not distinguish displacement of head relation and tail entities and therefore, it can not model anti-symmetric relations.

ComplEx~\cite{trouillon2016complex} solves the issue of DistMult by the idea that the complex conjugate of the tail makes it non-symmetric. By introducing complex-valued embeddings instead of real-valued embeddings to DistMult, the score of a triple in ComplEx is $Re(h^\top diag(r)\bar{t})$ with $\bar{t}$ the conjugate of t and $Re(.)$ is the real part of a complex value. ComplEx is not efficient in encoding composition rules~\cite{sun2019rotate}.
In RESCAL~\cite{nickel2011three} instead of a vector, a matrix represents the relation $r$, and performs outer products of $h$ and $t$ vectors to this matrix so that its score function becomes $h^\top R t$. A simplified version of RESCAL is HolE~\cite{nickel2016holographic} that defines a vector for $r$ and performs circular correlation of $h$ and $t$ has been found equivalent \cite{hayashi2017equivalence} to ComplEx. 
 
Another tensor factorization model is Canonical Polyadic (CP)~\cite{hitchcock1927expression}. In CP decomposition, each entity $e$ is represented by two vectors $h_e$, $t_e \in \mathbb{R}^d$, and each relation $r$ has a single embedding vector $ v_r \in \mathbb{R}^d $. MDE is similarly based on the idea of independent vector embeddings. A study~\cite{trouillon2017knowledge} suggests that in CP, the independence of vectors causes the poor performance of CP in KG completion, however, we show that the independent vectors can strengthen a model if they are combined complementarily.

SimplE~\cite{kazemi2018simple} analogous to CP, trains on two sets of subject and object entity vectors. SimplE's score function, $\frac{1}{2} \langle h_{e_i},r,t_{e_j} \rangle + \frac{1}{2} \langle h_{e_j},r^{-1},t_{e_j} \rangle$, is the average of two terms. The first term is similar to DistMult. However, its combination with the second term and using a second set of entity vectors allows SimplE to avoid the symmetric issue of DistMult. SimplE allows learning of symmetry, anti-symmetry and inversion patterns. However, it is unable to efficiently encode composition rules, since it does not model a bijection mapping from h to t through relation r.

In \textbf{Latent Distance Approaches} the score function is the distance between embedding vectors of entities and relations. 
In the view of social network analysis,~\cite{hoff2002latent} originally proposed distance of entities $-d(h, t)$ as the score function for modeling uni-relational graphs where $d(., .)$ means any arbitrary distance, such as Euclidean distance. SE~\cite{bordes2011learning} generalizes the distance for multi-relational data by incorporating a pair of relation matrices into it. TransE~\cite{bordes2013translating} represents relation and entities of a triple by a vector that has this relation
\begin{equation}\label{eq:1}
    S_1 = \parallel h+r-t \parallel_p
\end{equation}
where $\parallel . \parallel_p$ is the $p$-norm. To better distinguish entities with complex relations, TransH~\cite{wang2014knowledge} projects the vector of head and tail to a relation-specific hyperplane. 
Similarly, TransR follows the idea with relation-specific spaces and extends the distance function to $\parallel M_r h+r- M_r t \parallel_p$. RotatE~\cite{sun2019rotate} combines translation and rotation and defines the distance of a $t$ from tail $h$ which is rotated the amount $r$ as the score function of a triple $-d(h \circ r, t)$ where $\circ$ is Hadamard product.

\textbf{Neural Network Methods} train a neural network to learn the interaction of the \textbf{h}, \textbf{r} and \textbf{t}. ER-MLP \cite{dong2014knowledge} is a two layer feedforward neural network considering $h$, $r$ and $t$ vectors in the input. NTN~\cite{socher2013reasoning} is neural tensor network that concatenates head $h$ and tail $t$ vectors and feeds them to the first layer that has $r$ as weight. In another layer, it combines $h$ and $t$ with a tensor $R$ that represents $\textbf{r}$ and finally, for each relation, it defines an output layer $r$ to represent relation embeddings. In SME~\cite{bordes2014semantic} relation $r$ is once combined with the head $h$ to get $g_u(h, r)$, and similarly it is combined with the tail $t$ to get $g_v(t, r)$. SME defines a score function by the dot product of this two functions in the hidden layer. In the linear SME, $g(e,r)$ is equal to $M^1_u e+ M^2_u r + b_u$, and in the bilinear version, it is $M^1_u e \circ M^2_u r + b_u$. Here, $M$ refers to weight matrix and $b$ is a bias vector.

\section{MDE: Multiple Distance Embeddings}
\label{sec:MDE Multiple Distance Embeddings}
The score function of MDE involves multiple terms. We first explain the intuition behind each term and then explicate a framework that we suggest to efficiently utilize them such that we benefit from their strengths and avoid their weaknesses.

\textbf{Inverse Relation Learning:} Inverse relations can be a strong indicator in knowledge graphs. For example, if $IsParentOf(m, c)$ represents that a person $m$ is a parent of another person $c$, then this could imply $IsChildOf(c, m)$ assuming that this represents the person $c$ being the child of $m$. This indication is also valid in cases when this only holds in one direction, e.g. for the relations $IsMotherOf$ and $IsChildOf$. In such a case, even though the actual inverse $IsParentOf$ may not even exist in the KG, we can still benefit from inverse relation learning.  
To learn the inverse of the relations, we define a score function~S$_2$ :
\begin{equation}\label{eq:2}
 S_2 = \parallel t + r - h\parallel_p
 \end{equation}

\textbf{Symmetric Relations Learning:} It is possible to easily check that the formulation $\parallel h+r-t \parallel $ allows\footnote{We used the term ``it allows'' to imply that the encoding of such patterns do not inhibit the learning of relations having a particular pattern. Meanwhile in the literature SimplE uses ``it can encode'' and RotatE uses ``the model infers''.} learning of anti-symmetric pattern but when learning symmetric relations, $\parallel r \parallel$ tends toward zero which limits the ability of the model in separating entities specially if symmetric relations are frequent in the KG. 
For learning  symmetric relations, we suggest the term S$_3$ as a score function. It learns such relations more efficiently despite it is limited in the learning of antisymmetric relations. 

\begin{equation}\label{eq:3}
    S_3 = \parallel h+t-r \parallel_p
\end{equation}

\textbf{Lemma 1.} S$_1$ allows modeling antisymmetry, inversion and composition patterns and S$_2$ allows modeling symmetry patterns.
\begin{proof}
Let $r_1, r_2, r_3$ be relation vector representations and $e_i$, $e_j$, $e_k$ are entity representations. A relation $r_1$ between $(e_i, e_k)$ exists when a triple $(e_i , r_1 , e_k )$ exists and we show it by $r_1(e_i, e_k )$. Formally, we have the following results:
\begin{proof}[Antisymmetric Pattern]\let\qed\relax
If $r_1(e_i, e_j)$ and $r_1(e_j, e_i)$ hold, in equation \ref{eq:1} for S$_1$, then:
\begin{equation*}
 e_i + r_1 = e_j \quad \wedge \quad e_j + r_1 \not = e_i \quad  \Rightarrow \quad  e_i + 2 r_1 \not = e_i
 \qedhere
\end{equation*}
\end{proof}
\noindent
Thus S$_1$ allows encoding of relations with antisymmetric patterns.
\begin{proof}[Symmetric Pattern]\let\qed\relax
If $r_1(e_i, e_j)$ and $r_1(e_j, e_i)$ hold, for S$_3$ we have:
\begin{equation*}
  e_i +  e_j - r_1 = 0 \; \wedge \;  e_j +  e_i - r_1 = 0 \;  \Rightarrow \; e_j + e_i = r_1 \qedhere
\end{equation*}  
\end{proof}
\noindent
Therefore S$_3$ allows encoding relations with symmetric patterns. For S$_1$ we have:
\begin{proof}[Inversion Pattern]\let\qed\relax
If $r_1(e_i, e_j)$ and $r_2(e_j, e_i)$ hold, from Equation~\ref{eq:1} we have:
\begin{equation*}
e_i + r_1 =  e_j \quad \wedge \quad e_j + r_2 =  e_i \quad \Rightarrow  \quad  r_1 =  - r_2
\qedhere
\end{equation*}
\end{proof}
\noindent
Therefore S$_1$ allows encoding relations with inversion patterns.

\begin{proof}[Composition Pattern]\let\qed\relax
If $r_1(e_i, e_k)$ , $r_2(e_i, e_j)$ and, $r_3(e_j, e_k)$ hold, from equation \ref{eq:1} we have:
\begin{equation*}
e_i + r_1 = e_k \;  \wedge \; e_i + r_2 = e_j \; \wedge \; e_j + r_3 = e_k \;  \Rightarrow \;  r_2 + r_3 = r_1  \qedhere
\end{equation*}
\end{proof}
\noindent
Thus S$_1$ allows encoding relations with composition patterns.
\end{proof} 

\textbf{Relieving Limitations on Learning of Reflexive Relations:}\label{Relieving_Limitation_TransE}

A previous study~\cite{kazemi2018simple} highlighted the common limitations of TransE, FTransE, STransE, TransH and TransR for learning reflexive relations where these translation-based models force the reflexive relations to become symmetric and transitive. 
To relieve these limitations, we define S$_4$ as a score function which is similar to the score of RotatE i.e.,\ $\parallel h \circ r - t \parallel_p$ but with the Hadamard operation on the tail. In contrast to RotatE which represents entities as complex vectors, S$_4$ only holds in the real space:

\begin{equation}\label{eq:4}
    S_4 = \parallel h - r \circ t\parallel_p
\end{equation} 

\textbf{Lemma 2.} The following restrictions of translation based embeddings approaches do not apply to the S$_4$ score function. 
R1: if a relation $r$ is reflexive, on $\Delta \in \mathcal{E}$, $r$ it will be also symmetric on $\Delta$.
R2: if $r$ is reflexive on $\Delta \in \mathcal{E}$, $r$ it will be also be transitive on $\Delta$.  

\begin{proof}R1: For such reflexive $r_1$, if $r_1(e_i, e_i)$ then $r_l(e_j, e_j)$. 
In this equation we have:
$$e_i = r_1 e_i \wedge e_j = r_1 e_j \Rightarrow r_1 = U \not\Rightarrow e_i = r_1 e_j$$
where $U$ is unit tensor.

R2: For such reflexive $r_1$, if $r_1(e_i, e_j)$ and $r_l(e_j, e_k)$ then $r_1(e_j, e_i)$ and $r_l(e_k, e_j)$. 
In the above equation we have:
\begin{equation*}
\begin{split}
    e_i = r_1 e_j  \wedge e_j = r_1 e_k  & \Rightarrow  e_i = r_1 r_1 e_j e_k \wedge  r_i = U  \\
    & \Rightarrow e_i = e_j e_k \\
    & \not\Rightarrow e_i + e_k = r_l
\end{split}
\end{equation*}
\qedhere
\end{proof}

\textbf{Model Definition:} To incorporate different views to the relations between entities, we define these settings for the model:
\begin{enumerate}
    \item Using limit-based loss instead of margin ranking loss.
    \item Each aggregated term in the score represents a different view of entities and relations with an independent set of embedding vectors.
    \item In contrast to ensemble approaches that incorporate models by training independently and testing them together, MDE is based on multi-objective optimization~\cite{marler2004survey} that jointly minimizes the objective functions. 
\end{enumerate}

However, when aggregating different terms in the score function, the summation of opposite vectors can cause the norm of these vectors to diminish during the optimization. For example if S$_1$ and S$_3$ are added together, the minimization would lead to relation(r) vectors with zero norm value.
To address this issue, we represent the same entities with independent variables in different distance functions. 


Based on CP, MDE considers four vectors $e_i, e_j, e_k, e_l,  \in \mathbbm{R}^d$ as the embedding vector of each entity $\textbf{e}$
, and four vectors $r_i, r_j, r_k, r_l \in \mathbbm{R}^d$ for each relation $\textbf{r}$. 

The score function of MDE for a triple $(\textbf{h}$, $ \textbf{r}$, $\textbf{t})$ is defined as weighted sum of listed score functions:


\begin{equation}\label{eq:5}
f_{MDE} = w_1 S_1^i~+~ w_2 S_2^j~+~ w_3 S_3^k~+~w_4S_4^l~ - \psi 
\end{equation}

\begin{figure*}[t]
\centering
\includegraphics[width=0.75\textwidth]{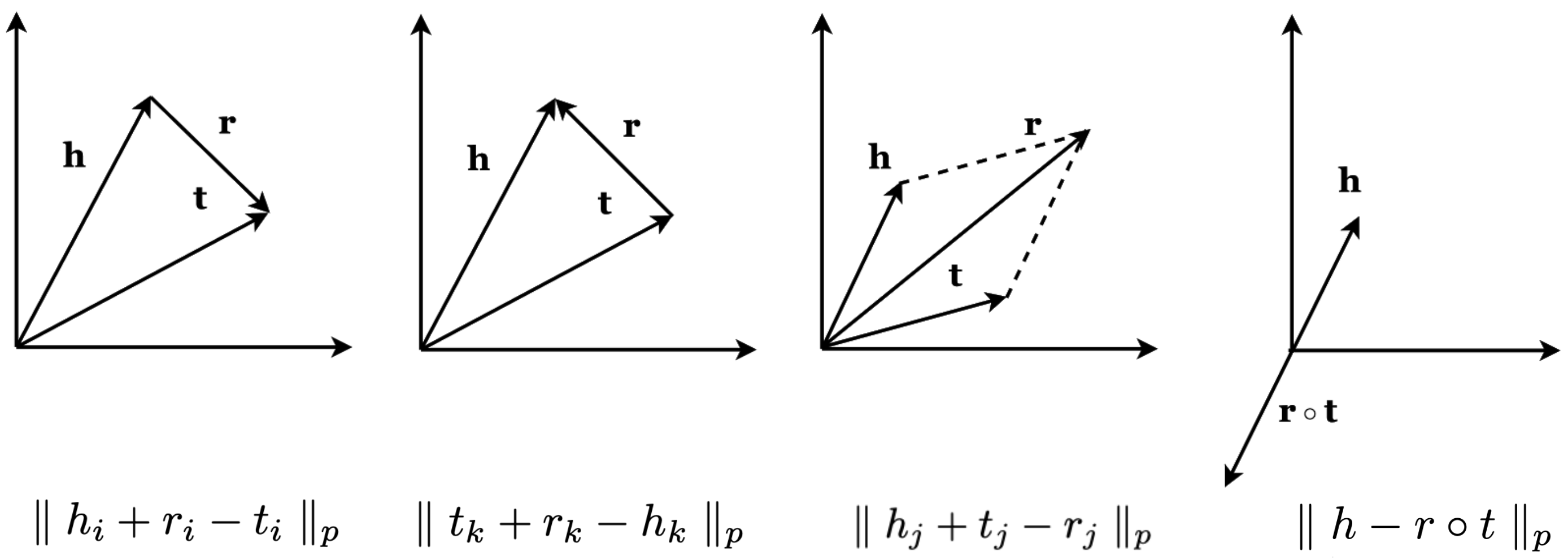}
\caption{Geometric illustration of the translation terms considered in MDE.}
\label{fig:mdediagram}
\end{figure*}

where $\psi, w_1, w_2, w_3, w_4 \in \mathbbm{R}$ are constant values. Figure~\ref{fig:mdediagram} displays the geometric illustration of the four translation terms considered in MDE. In the following, we show using $\psi$ and limit-based loss, the combination of the terms in Equation~\eqref{eq:5} is efficient, such that if one of the terms recognises if a sample is true $F_{MDE}$ would also recognize it.

\textbf{Limit-based Loss:}
Because margin ranking loss minimizes the sum of error from directly comparing the score of negative to positive samples, when applying it to translation embeddings, it is possible that the score of a correct triplet is not small enough to hold the relation of the score function~\cite{zhou2017learning}. To enforce the scores of positive triples become lower than those of negative ones, \cite{zhou2017learning} defines limited-based loss which minimizes the objective function such that the score for all the positive samples become less than a fixed limit. \cite{sun2018bootstrapping} extends the limit-based loss so that the score of the negative samples become greater than a fixed limit. We train our model with the same loss function which is:

\begin{equation}\label{eq:6}
loss =  \beta_1 \sum_{\tau\in \mathds{T}^+} [f(\tau)- \gamma_1]_+ + \beta_2 \sum_{\tau'\in \mathds{T}^-} [\gamma_2 - f(\tau')]_+ 
\end{equation}

where $[.]_+\, =\, \max(., 0)$,\,$\gamma_1, \gamma_2 \in \mathbbm{R}^+$. $\mathds{T}^+ ,\mathds{T}^-$ are the sets of positive and negative samples and $\beta_1, \beta_2 >0$ are constants denoting the importance of the positive and negative samples. This version of limit-based loss minimizes the aggregated error such that the score for the positive samples becomes less than $\gamma_1$ and the score for negative samples becomes greater than $\gamma_2$. To find the optimal limits for the limit-based loss, we suggest updating the limits during the training.

\paragraph*{Time Complexity and Parameter Growth:}
Considering the ever growth of KGs and the expansion of the web, it is crucial that the time and memory complexity of a relational mode be minimal. Despite the limitations in expressivity, TransE is one of the popular models on large datasets due to its scalability. With $O(d)$ time complexity (of one mini-batch), where $d$ is the size of embedding vectors, it is more efficient than RESCAL, NTN, and the neural network models. 
Similar to TransE, the time complexity of MDE is $O(d)$. Due to the additive construction of MDE, the inclusion of more distance terms keeps the time complexity linear in the size of vector embeddings.

\section{Model Extensions}\label{sec:extensions}
\subsection{Searching for the limits in the limit-based Loss}
While the limit-based loss resolves the issue of margin ranking loss with distance based embeddings, it does not provide a way to find the optimal limits. Therefore the mechanism to find limits for each dataset and hyper-parameter is the try and error. To address this issue, we suggest updating the limits in the limit-based loss function during the training iterations. We denote the moving-limit loss by $loss_{guide}$.

\begin{equation}\label{eq:5.5}
\begin{split}
loss_{guide} = \lim_{\delta ,\delta' \to \gamma_1} & \beta_1 \sum_{\tau\in \mathds{T}^+} [f(\tau)- (\gamma_1 - \delta)]_+ \\
& + \beta_2 \sum_{\tau'\in \mathds{T}^-} [(\gamma_2 - \delta') - f(\tau')]_+ 
\end{split}
\end{equation}
where the initial value of $\delta, \delta'$ is $0$.  In this formulation, we increase the $\delta, \delta'$ toward $\gamma_1$ and $\gamma_2$ during the training iterations such that the error for positive samples minimizes as much as possible.
We test on the validation set after each 50 epoch and take those limits that give the best value during the tests. The details of the search for limits is explained in the algorithm below. 
After observing the most promising values for limits in the preset number of iterations, we stop the search and perform the training while having the $\delta$ values fixed(fixed limit-base loss) to allow the adaptive learning to reach loss values smaller than the $threshold$.


We based this approach on the idea of adaptive learning rate~\cite{zeiler2012adadelta}, where the Adadelta optimizer adapts the learning rate after each iteration, therefore in the $loss_{guided}$ we can update the limits without stopping the training iterations. In our experiments, the variables in the algorithm, are as follows: $\delta{_0}=0$, $threshold =$ 0.05, $\xi =$ 0.1. 
	\begin{algorithmic}[1]
		\State {{\bf Initialize:} $\delta=\delta'=\delta{_0} , ~ \gamma_1 = \gamma_2 \in \mathbbm{R^+} , ~ \psi \in \mathbbm{R}$}
		\State {{\bf Initialize:} $i = 0, ~ \xi \in \mathbbm{R^+}, ~ threshold \in \mathbbm{R^+}$ }
		\State {{Inside training iterations:}}
		\If{Using $loss_{guided}$ instead of $loss_{limit-based}$}
		    \State {{$loss^+ = \beta_1 \sum_{\tau\in \mathds{T}^+} [f(\tau)- (\gamma_1 - \delta)]_+$}}
    		\State {{$loss^- = \beta_2 \sum_{\tau'\in \mathds{T}^-} [(\gamma_2 - \delta') - f(\tau')]_+$}}
	    	\State {{$loss ~=~ loss^+ ~+~ loss^-$}}
		    \If{ $loss^+ = 0 \And \gamma_1 \ge \xi$} 
		        \State {{$\delta = \delta +  \xi$}}
	        	\If{ $loss^- > threshold \And \gamma_2 \ge \xi$} 
	        	\State{{$\delta' = \delta' +  \xi$}}
			    \EndIf
		    \EndIf
	    \EndIf
	    \If{Using $loss_{limit-based}$}
	        \State{$loss$ = the result from Equation \eqref{eq:6}}
	    \EndIf
	\end{algorithmic}



%

\textbf{Lemma 3.} There exist $\psi$ and $\gamma_1, \gamma_2 \ge 0$ ($\gamma_1 \ge \gamma_2$), such that only if one of the terms in $f_{MDE}$ estimates a fact as true, $f_{MDE}$ also predicts it as a true fact. Consequently, the same also holds for the capability of MDE to allow learning of different relation patterns.

\begin{proof} We show there are boundaries for $\gamma_1, \gamma_2, w_1, w_2, w_3, w_4$, such that learning a fact by one of the terms in $f_{MDE}$ is enough to classify a fact correctly. 

The case to prove is when three of the distance functions classify a fact negative $N$ and the one distance function e.g.\ S$_2$ classify it as positive $P$, and the case that S$_1$ and S$_3$ classify a fact as positive and S$_2$ classify it as negative. We set $w_1 = w_3 = 1/4$ and  $w_2  = 1/2$ and assume that S$um$ is the value estimated by the score function of MDE, we have:
\begin{equation}\label{eq:8}
a > \frac{N}{2} \ge \frac{\gamma_2}{2} \wedge \frac{\gamma_1}{2} > \frac{P}{2} \ge 0  \Rightarrow a + \frac{\gamma_1}{2} > Sum + \psi \ge \frac{\gamma_2}{2}  
\end{equation}

There exist $a= 2$ and $\gamma_1 = \gamma_2= 2$ and $\psi = 1$ that satisfy $\gamma_1 > Sum  \ge 0 $ and the inequality \ref{eq:8}.
\end{proof}
It is notable that without the introduction of $\psi$ and the limits $\gamma_1, \gamma_2$ from the limit-based loss, Lemma 3 does not hold and framing the model with this settings makes the efficient combination of the terms in $f_{MDE}$ possible. In case that future studies discover new interesting distances, this Lemma shows how to basically integrate them into MDE.

In contrast to SimplE that ties the relation vectors of two terms in the score together, MDE does not directly relate them to take advantage of the independent relation and entity vectors in combining opposite terms.

The learning of the symmetric relations is previously studied (e.g.\ in \cite{yang2014embedding,sun2019rotate}) and \cite{lin2015modeling} studied the training over the inverse of relations, however providing a way to gather all these benefits in one model is a novelty of MDE. Besides, complementary modeling of different vector-based views of a knowledge graph is a novel contribution. 

\begin{figure}[t]
\centering
\includegraphics[width=0.26\textwidth]{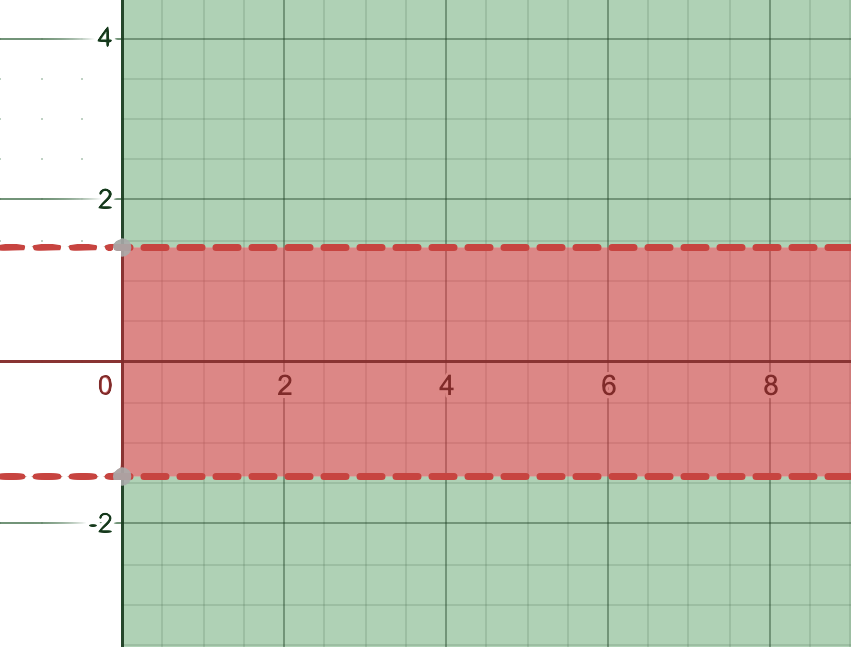}
\caption{Illustration of the possible positioning of score  values for MDE$_{NN}$ on WN18RR where the value of $\gamma_1$ and $\gamma_2$ is 2.}
\label{fig:mdennloss}
\end{figure}

\subsection{MDE$_{NN}$: MDE as a Neural Network}

The score of MDE is already aggregating a multiplication of vectors to weights. We take advantage of this setting to model MDE as a layer of a neural network that allows learning the embedding vectors and multiplied weights jointly during the optimization. To create such a neural network we multiply $\psi$ by a weight $w_5$ and we feed the MDE score to an activation function. We call this extension of MDE as MDE$_{NN}$:

\begin{equation}\label{eq:9}
\begin{split}
f_{MDE_{NN}} = F( & \parallel w_1 S_1^i\parallel_p + \parallel w_2 S_2^j\parallel_p + \parallel w_3 S_3^k\parallel_p \\&+ \parallel w_4S_4^l\parallel_p~ + \parallel w_5 \parallel_p c - \psi ) 
\end{split}
\end{equation}
where $F$ is $Tanhshrink$ activation function with the formulation 
\begin{equation}\label{eq:10}
Tanhshrink(x)= x - Tanh(x) 
\end{equation}
and $w_1,~w_2,\dots,~w_5$ are elements of the latent vector $w$ that are estimated during the training of the model and $c$ and $\psi$ are constants. Similarly we add $y$ and $z$ as latent vectors multiplied to the first and the second elements in the Equations~\ref{eq:1},~\ref{eq:2},~\ref{eq:3}~\&~\ref{eq:4}. For example S$_1$ in MDE$_{NN}$ becomes:

\begin{equation}\label{eq:11}
    S_1 = \parallel y_1 h+ z_1 r - t \parallel_p
\end{equation}

This framing of MDE reduces the number of hyper parameters. In addition, the major advantage of MDE$_{NN}$ --in comparison to the linear combination of terms in MDE-- is that the $Tanhshrink$ activation function allows the non-linear mappings between the embedding vectors and the expected target values for the loss function over positive and the negative samples. 

Since $Tanhshrink$ has a range of $\mathbb{R}$ it allows setting large values for $\gamma_1$ and $\gamma_2$. For example for WN18RR we set their value to 1.9. It is notable that the classic activation functions such as $sigmoid$ and $Tanh$ are not suitable to be used as activation functions here because they cannot converge the loss function to limit values larger than one.

To generate a non-linear loss function for MDE$_{NN}$, we combine the square of positive loss and the negative loss values:
\begin{equation}\label{eq:12}
\begin{split}
loss_{{MDE}_{NN}}  &=  ( \sum_{\tau\in \mathds{T}^+} [f(\tau)- \gamma_1]_+  )^2\\
&+   ( \sum_{\tau'\in \mathds{T}^-} [\gamma_2 - f(\tau')]_+ )^2\\
\end{split}
\end{equation}

Figure~\ref{fig:mdennloss} shows the positioning of the score values for MDE$_{NN}$ on WN18RR in which $\gamma_1$ and $\gamma_2$ is 2. The horizontal axis indicates the sample numbers and the vertical axis indicates their loss values. The score values for negative samples, $f(\tau')$ lay on the green area and score values for the positive samples, $f(\tau)$ lay on the red area.

\section{Experiments}\label{sec:experiments}
\textbf{Datasets:}
We experimented on four standard datasets: WN18 and FB15k which were extracted by Bordes \textit{et al.} in~\cite{bordes2013translating} from Wordnet~\cite{miller1995wordnet} and Freebase~\cite{bollacker2008freebase} respectively. We used the same train/valid/test sets as in~\cite{bordes2013translating}. WN18 contains 40\,943 entities, 18 relations and 141\,442 train triples. FB15k contains 14\,951 entities, 1\,345 relations and 483\,142 train triples. 
In order to test the expressiveness ability rather than relational pattern learning power of models, FB15k-237~\cite{toutanova2015observed} and WN18RR~\cite{dettmers2018convolutional} exclude the triples with inverse relations from FB15k and WN18 which reduced the size of their training data to 56\% and 61\% respectively. Table~\ref{datasetinfo} summarizes the statistics of these knowledge graphs.  

\begin{table} 
\centering
    \begin{tabular}{c|c|c|c|c|c}
       Dataset & \#entity & \#relation & \#training &  \#validation & \#test   \\
       \hline
       FB15k & 14\,951 & 1\,345 & 483\,142 & 50\,000 & 59\,071 \\
       \rowcolor{LightGray}WN18 & 40\,943& 18 & 141\,442 & 5\,000 & 5\,000 \\
       FB15k-237 & 14\,541 & 237 & 272\,115 & 17\,535 & 20\,466 \\
       \rowcolor{LightGray}WN18RR & 40\,943 & 11 & 86\,835 & 3\,034 & 3\,134 \\
    \end{tabular}
    \caption{Number of entities, relations, and triples in each division.}
    \label{datasetinfo} 
\end{table}

\textbf{Baselines:} 
We compare MDE with several state-of-the-art relational learning approaches. Our baselines include TransE, RESCAL, DistMult, NTN, ER-MLP, ComplEx and SimplE. We report the results of TransE, DistMult, and ComplEx from~\cite{trouillon2016complex} and the results of TransR and NTN from~\cite{nguyen2017overview}, and ER-MLP from~\cite{nickel2016holographic}. The results on the inverse relation excluded datasets are from the Table13 of~\cite{sun2019rotate} for both TransE and RotatE. And the rest are from~\cite{dettmers2018convolutional}\footnote{Scores of ConvE on FB15k is from \url{https://github.com/TimDettmers/ConvE/issues/26}}.

\textbf{Evaluation Settings:} 
We evaluate the link prediction performance by ranking the score of each test triple against its versions with replaced head, and once for tail. Then we compute the hit at N (Hit@N), mean rank (MR) and mean reciprocal rank (MRR) of these rankings. We report the evaluations in the filtered setting. 
\begin{figure*}[t]
 \centering
  \subcaptionbox{Positive triple including Similar$\_$to\label{fig3:a}}{\includegraphics[width=1.6in]{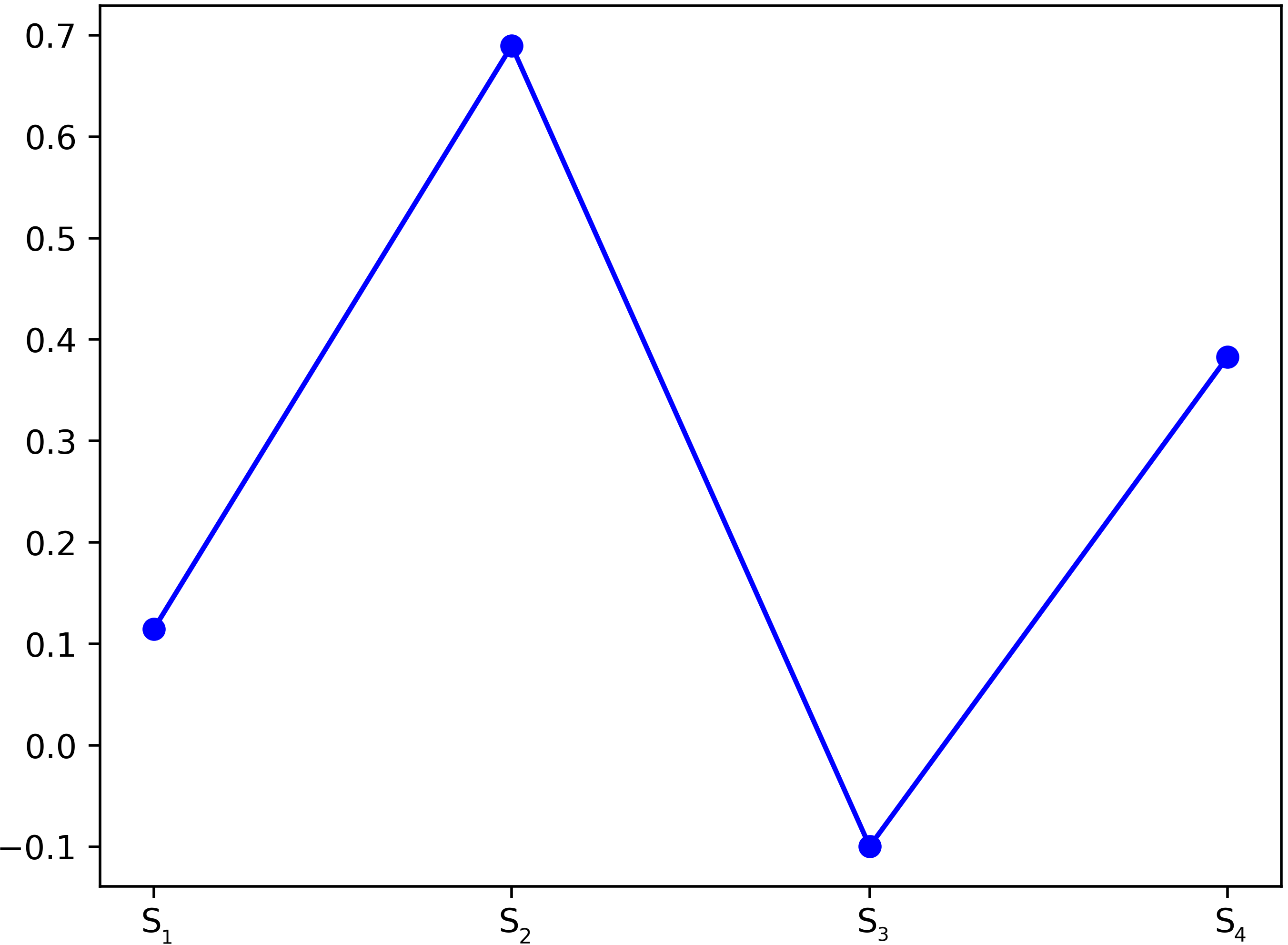}}\hspace{1em}%
  \subcaptionbox{Negative triple including symmetric pattern Similar$\_$to\label{fig3:b}}{\includegraphics[width=1.6in]{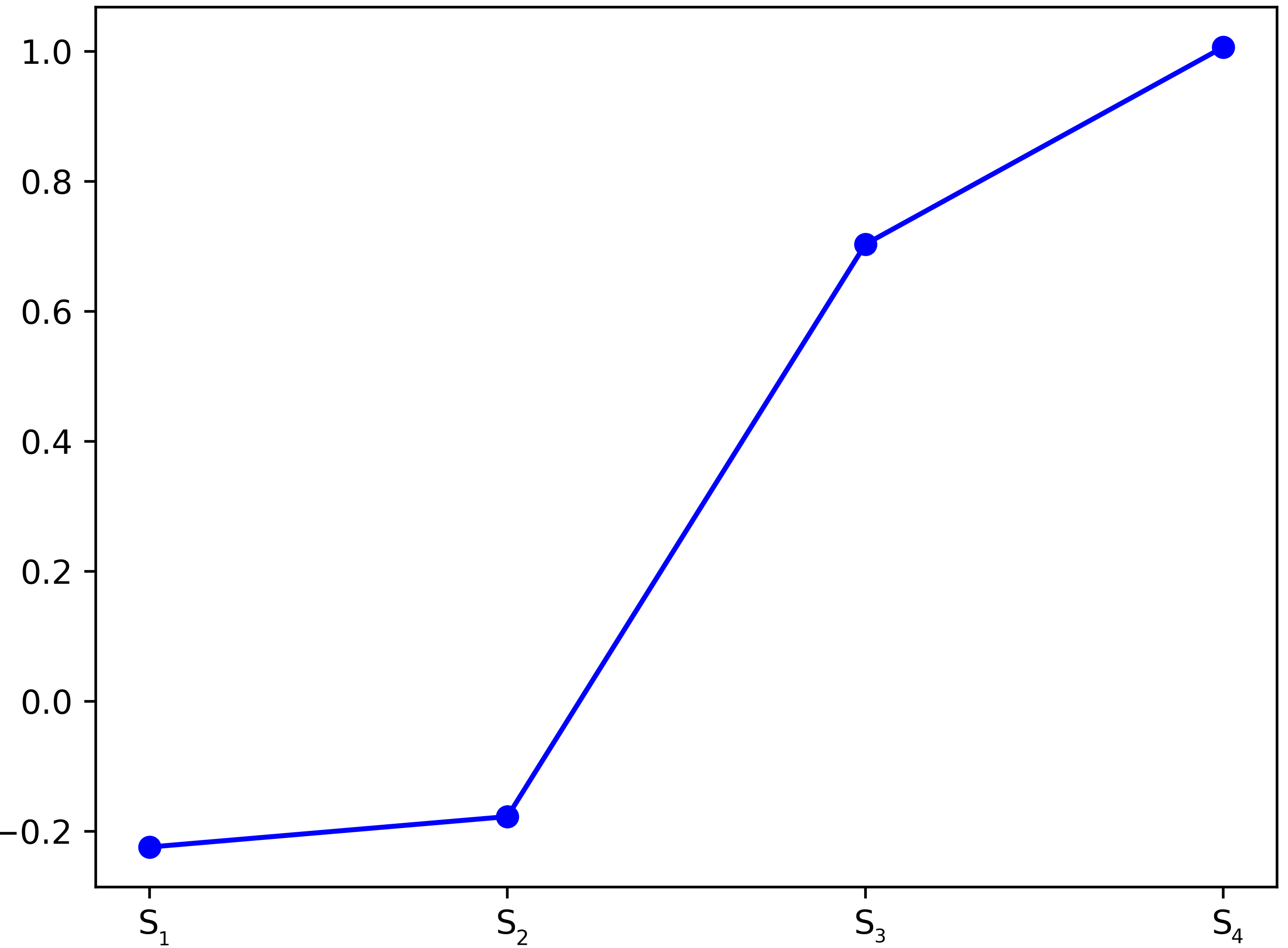}}
 \subcaptionbox{hypernym + hyponym\label{fig3:c}}{\includegraphics[width=1.6in]{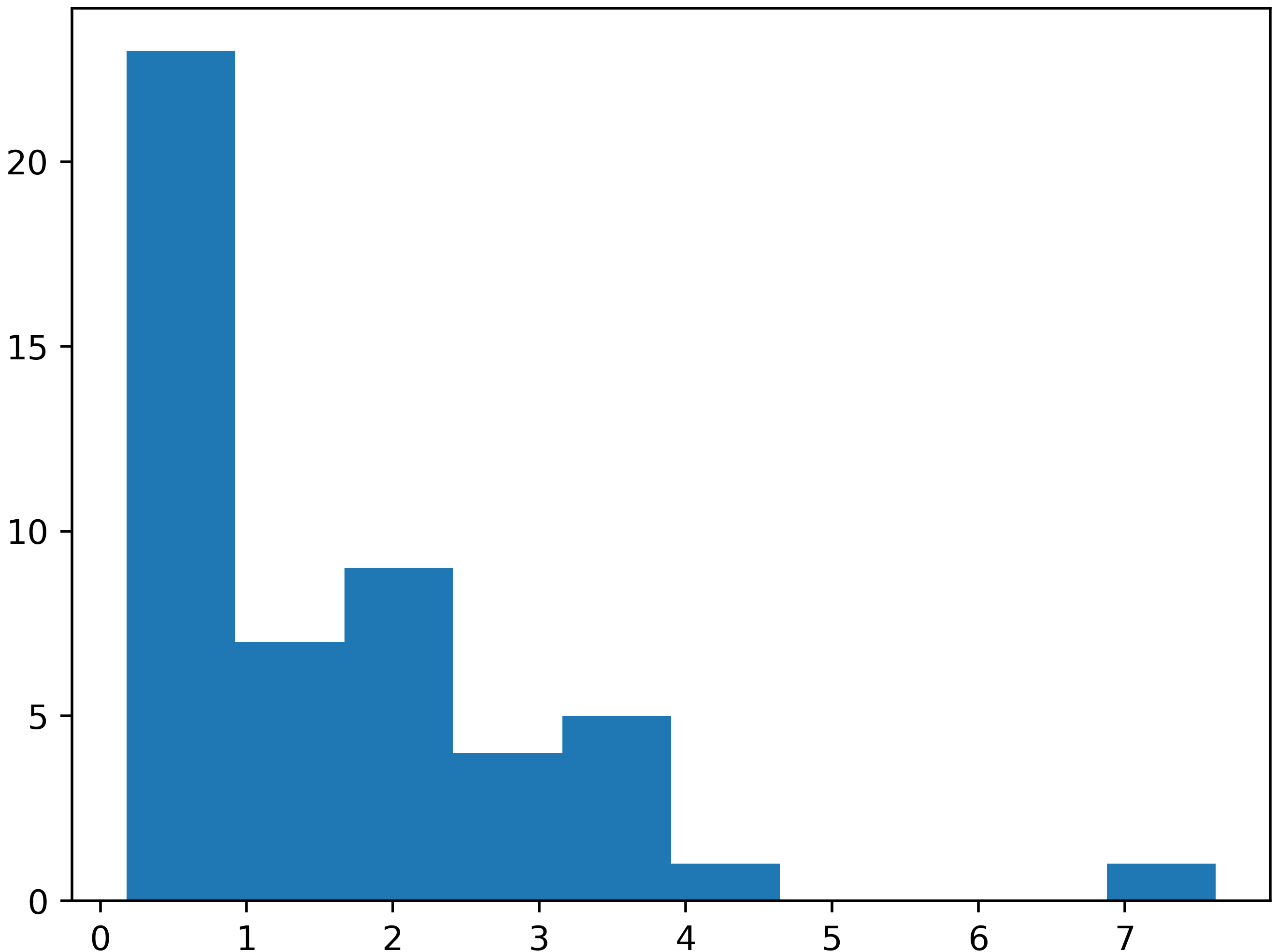}}
    
\subcaptionbox{r$_1$\label{fig3:d}}{\includegraphics[width=1.27in]{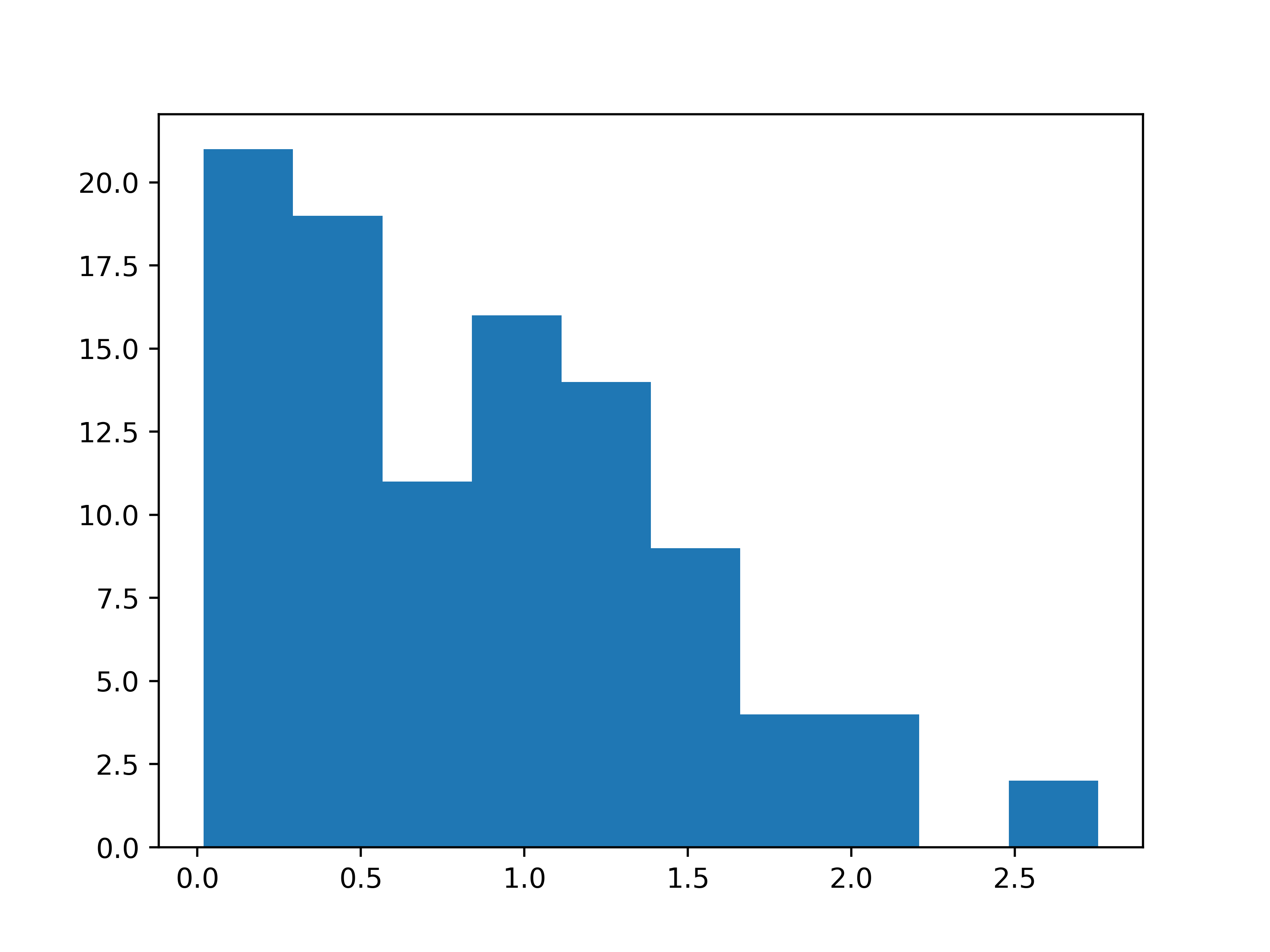}}%
\subcaptionbox{r$_2$\label{fig3:e}}{\includegraphics[width=1.27in]{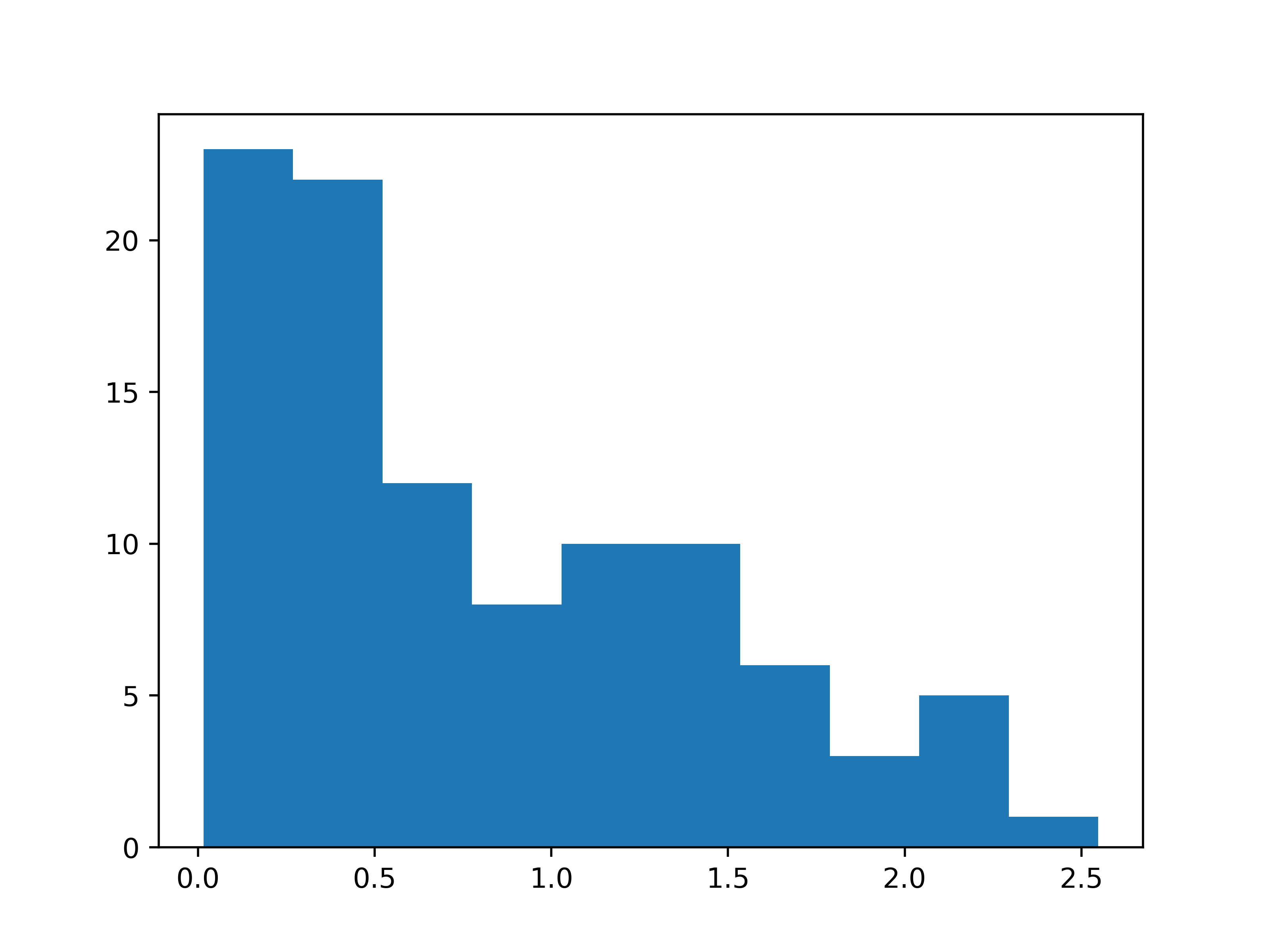}}%
\subcaptionbox{r$_3$\label{fig3:f}}{\includegraphics[width=1.27in]{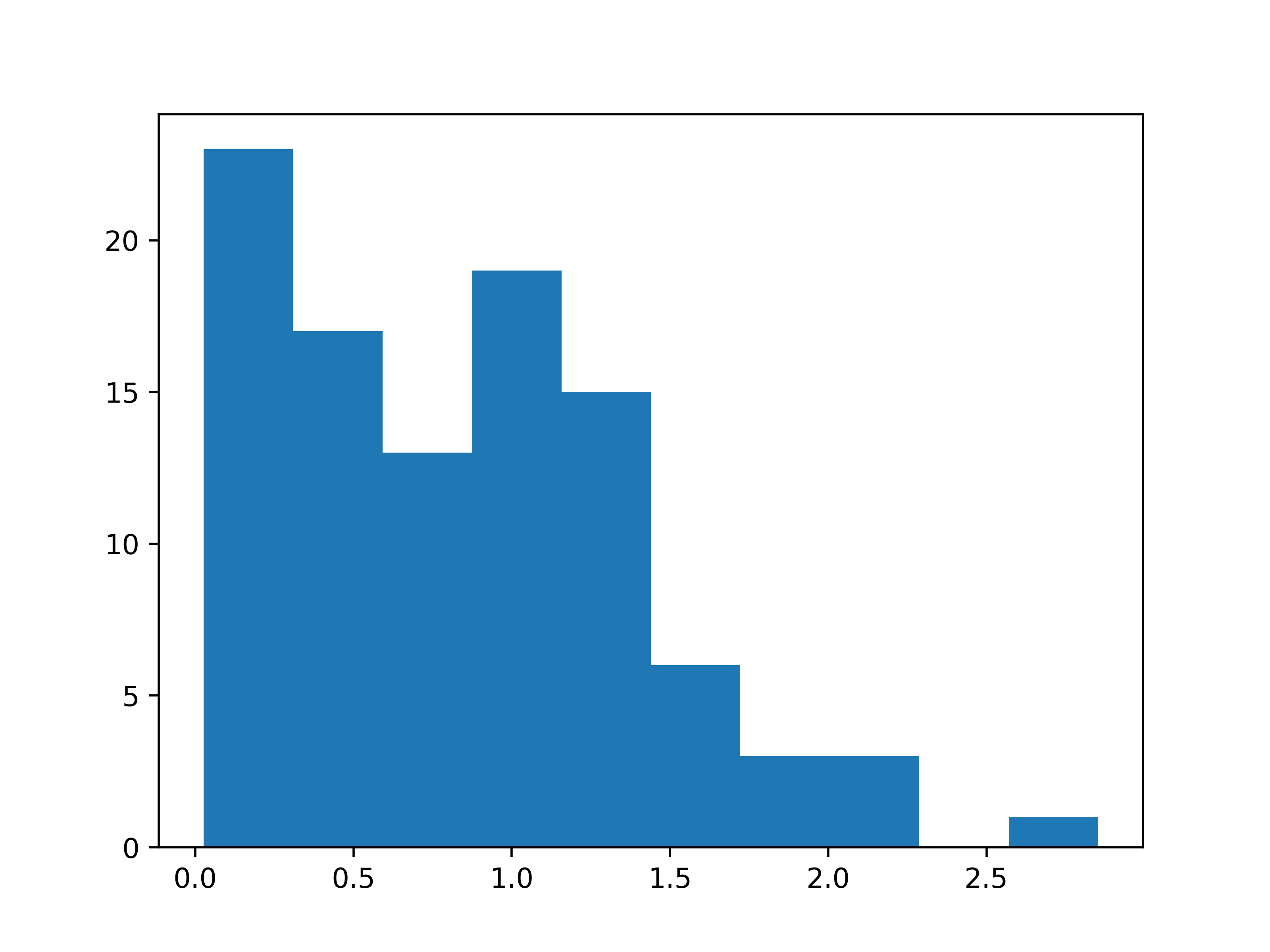}}%
\subcaptionbox{r$_1$+r$_2$-r$_3$\label{fig3:g}}{\includegraphics[width=1.27in]{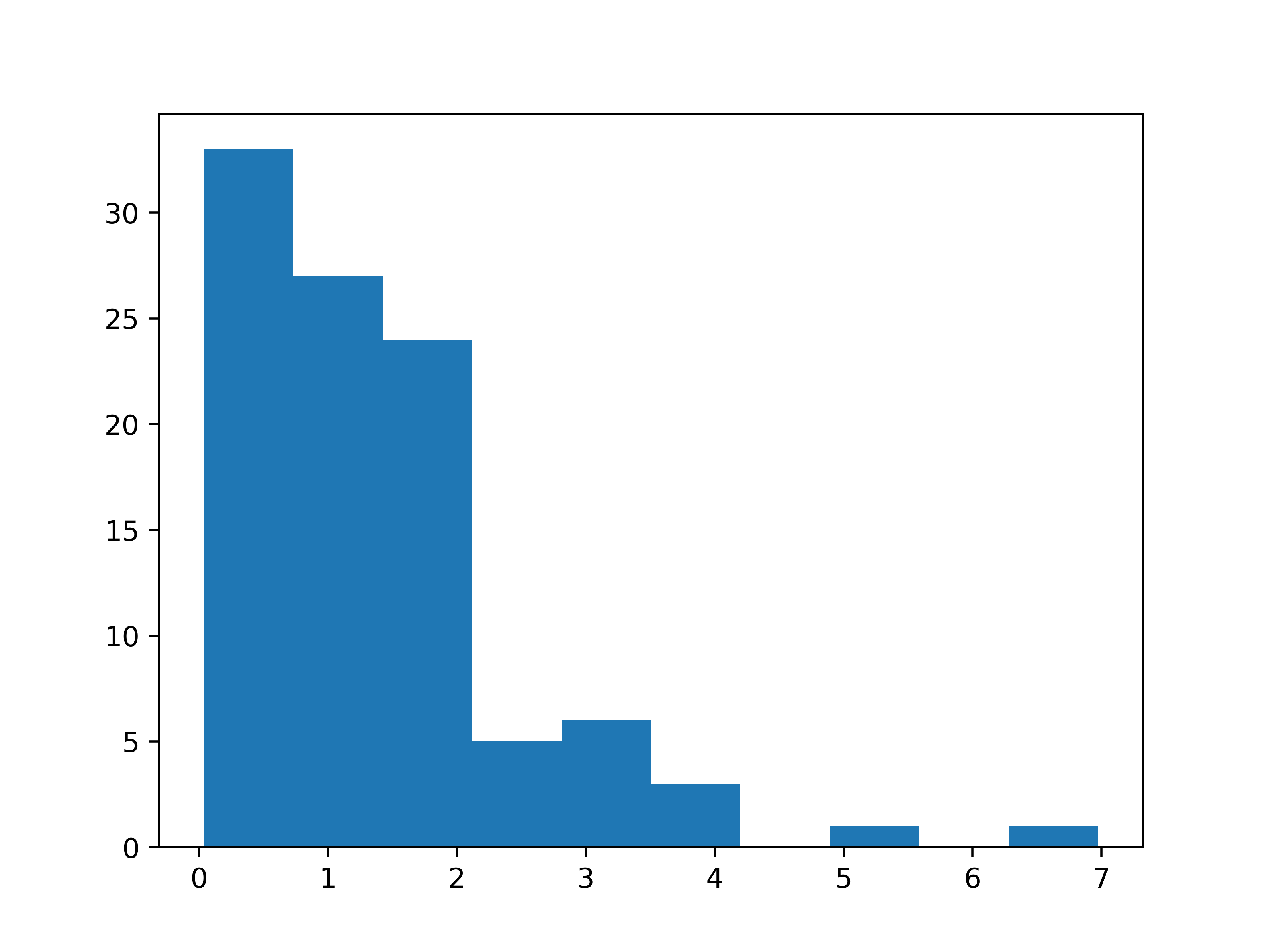}}

    \caption{Prediction of each term in MDE score for a symmetric relation in a positive triple in Figure (a) and its corrupted version with the same head and tail in Figure (b). Lower values indicate that a triple is recognized as positive. Figure (c) shows the histogram diagram of the elements of two the sum of two inverse relations, hypernym and hyponym in S$_1$. Figures~(d,\,e,\,f\,\&\,g) show the norm of the elements in vectors r$_1$, r$_2$, r$_3$ and r$_1$+r$_2$-r$_3$ where r$_3$ is composed of r$_1$ and r$_2$, where r$_1$ represents~\footnotesize/award/award\_category/nominees./award/award\_nominatio/nominated\_for \normalsize and r$_2$ represents  \footnotesize/award/award\_nominee/award\_nominations./award /award\_nomination/nominated\_for \normalsize and r$_3$ represents  \footnotesize /award/award\_winner/awards\_won./award/award\_honor/award\_winner\normalsize.} 
\label{fig:relationPatterns}
\end{figure*}

\textbf{Implementation:}
We implemented MDE in PyTorch\footnote{\url{https://pytorch.org}}.  Following~\cite{bordes2011learning}, we generated one negative example per positive example for all the datasets. We used Adadelta~\cite{zeiler2012adadelta} as the optimizer and fine-tuned the hyperparameters on the validation dataset. The ranges of the hyperparameters are set as follows: embedding dimension 25, 50, 100, 200, batch size in range of 1024 to 1725 and iterations 50, 100, 1000, 1500, 2500, 3600. We set the initial learning rate on all datasets to 10. 
For MDE, the best embedding size and $\gamma_1$ and $\gamma_2$ and $\beta_1$ and $\beta_2$ values on WN18 were 50 and 1.9, 1.9, 2 and 1 respectively and for FB15k were 200, 10, 13, 1, 1. The best found embedding size and $\gamma_1$ and $\gamma_2$ and $\beta_1$ and $\beta_2$ values on FB15k-237 were 100, 9, 9, 1 and 1 respectively and for WN18RR were 50, 2, 2, 5 and 1. 

We selected the coefficient of terms in~\eqref{eq:5}, by grid search, with the condition that they make a convex combination, in the range 0.1 to 1.0 and testing those combinations of the coefficients where they create a convex combination. Found values are $w_1$ = 0.16, $w_2$ = 0.33, $w_3$ = 0.16, $w_4$=0.33. We also tested for the best value for $\psi$ between $\{0.1, 0.2, $\dots$, 1.5\}$. We use $\psi$ = 1.2 for the MDE experiments. We use the value 2 for p in p-norm through the paper. To regulate the loss function and to avoid over-fitting, we estimate the score function for two sets of independent vectors and we take their average in the prediction. Another advantage of this operation is the reduction of required training iterations. 

For WN18RR experiment of MDE$_{NN}$, we use the same parameters as in MDE for $\gamma_1$, $\gamma_2$ and the embedding size.
We use adaptive learning rate method for both MDE and MDE$_{NN}$ in our experiments. 

The current framework of KG embedding model evaluations is based on the open-world assumption where the generation of an unlimited number of negative samples is possible. In this setting, it becomes debatable to consider negative sample generation as a part of the model since it significantly influences the ranking results.
In particular, RotatE efficiently assimilates the effect of many negative samples in self-adversarial negative sampling technique.
We verify the influence of this sampling method on the MDE results and to distinguish it we call this implementation MDE$_{adv}$. For this implementation, we use Adam as the optimizer similar to RotatE. We select dimension 400, learning rate 0.0005, batch size 512 and 624 negative samples per positive sample for the test on WN18RR. For FB15k-237, we test the model with dimension 1000, learning rate 0.0005, the batch size 240 and 1224 negative samples per positive sample.

\subsection{Relation Pattern Implicit Inference}
To verify the implicit learning of relation patterns, we evaluate our model on Countries dataset~\cite{bouchard2015approximate,nickel2016holographic}. This dataset is curated in order to explicitly assess the ability of the link prediction models for composition pattern modeling and implicit inference. It is made from 2 relations and 272 entities, where the entities include 244 countries, 5 regions and 23 subregions. In comparison to general link prediction tasks on knowledge graphs, evaluation queries in Countries are specified only to the form locatedIn(c, ?), where, the answer is one of the five regions. The Countries dataset is made of 3 tasks, and each one requires inferring a composition pattern with increasing length and difficulty. The measure for this evaluation is usually AUC-PR.

\begin{table}
\centering
\resizebox{\columnwidth}{!}{
    \begin{tabular}{c|ccc}
        & &\textbf{Countries}(AUC-PR) & \\
        Model & S1 & S2 & S3 \\
        \hline
    DistMult~\cite{dismult} & \textbf{1.00} $\pm$ \textbf{0.00} & 0.72 $\pm$ 0.12 & 0.52 $\pm$ 0.07 \\
    \rowcolor{LightGray}ComplEx~\cite{trouillon2016complex} & 0.97 $\pm$ 0.02 & 0.57 $\pm$ 0.10 & 0.43 $\pm$ 0.07  \\
    ConvE~\cite{dettmers2018convolutional} & \textbf{1.00} $\pm$ \textbf{0.00} & 0.99 $\pm$ 0.01 & 0.86 $\pm$ 0.05 \\
    \rowcolor{LightGray}RotatE~\cite{sun2019rotate} &  \textbf{1.00} $\pm$ \textbf{0.00} & \textbf{1.00} $\pm$ \textbf{0.00} & 0.95 $\pm$ 0.00 \\
    MDE & \textbf{1.00} $\pm$ \textbf{0.00} & \textbf{1.00} $\pm$ \textbf{0.00} & \textbf{1.00} $\pm$ \textbf{0.00} 
    \\
    \end{tabular}
    }
    \caption{Results on the Countries datasets. Results of RotatE are taken from~\cite{sun2019rotate} and the results of the other models are from~\cite{dettmers2018convolutional}.}.
\label{tab:counties}
\end{table}%

Table~\ref{tab:counties}, shows that our model performs significantly better than the previous models. While RotatE outperforms older models on S1 and S2, MDE gains the best result on S1 and S2 as well as S3, which is the most difficult task.
We also evaluate if MDE embeddings implicitly represent different relation patters.

\textbf{Symmetry} pattern requires S$_3$ term to correctly distinguish positive and negative samples for MDE. We investigate the relation embeddings from a 50-dimensional MDE trained on WN18. Figure~\ref{fig3:a} gives the value of different terms for a triple with symmetric relation ``similar$\_$to'' between the entities ``pointed'' and ``sharpened''. Since the smaller score values of MDE are suggesting that a triple is a positive sample, the smaller values of individual terms in the model would also influence the overall model to recognize a triple as positive. S$_3$ shows the smallest value between all the terms.
Figure~\ref{fig3:b} illustrates the values of terms for the negative sample (pointed, similar$\_$to, pointed) where S$_1$ and S$_2$ scores are low due to their incapability in recognizing a negative sample when the head and tail are the same. However, S$_3$ adjusts the overall MDE score by producing a great number that compensates the low S$_1$ and S$_2$ results.

\textbf{Inversion} pattern requires inverse relations in S$_1$ and S$_2$ terms to have inverse angles. Figure~\ref{fig3:c} shows the histogram of the elements of the sum of hypernym and hyponym relations in S$_1$. We can see from this Figure that most of the elements in this two relations have opposite values.

\textbf{Composition} pattern requires the embedding vectors of the composed relation to be the addition of the other two relations in S$_1$. We train a 200-dimensional MDE model to verify the implicit inference of the composition patterns on FB15k-237. Figure~\ref{fig3:d} to~\ref{fig3:g} illustrate that most of the elements in  r$_1$ + r$_2$ - r$_3$ are near zero where r$_3$ is composed of r$_1$ and r$_2$ relations.

\begin{table} 
\centering
\resizebox{\columnwidth}{!}{
    \begin{tabular}{c|ccc|cccc}
        & & \textbf{WN18} & & & \textbf{FB15k} & \\
        Model & MR & MRR & Hit@10 & MR & MRR & Hit@10 \\
        \hline
        \rowcolor{LightGray}TransE \cite{bordes2013translating}& -- & 0.454 & 0.934 & -- & 0.380 & 0.641 \\
        TransH \cite{wang2014knowledgeTransH}& 303 & -- & 0.867 & 87 & -- & 0.644 \\
        \rowcolor{LightGray}STransE \cite{StransE} & 206 & 0.657 & 0.934 & 69 & 0.543 & 0.797\\
        RESCAL~\cite{nickel2011three} & -- & 0.890 & 0.928 & -- & 0.354 & 0.587\\  
        \rowcolor{LightGray}DistMult~\cite{dismult} & --  & 0.822 & 0.936 & -- &0.654 & 0.824 \\
        SimplE ~\cite{kazemi2018simple} & -- & 0.942 & 0.947 & -- & 0.727 & 0.838 \\
        \rowcolor{LightGray}NTN\cite{socher2013reasoning} & -- & 0.53 & 0.661 & -- & 0.25 &  0.414 \\
        ER-MLP~\cite{dong2014knowledge} & -- & 0.712 & 0.863 & -- & 0.288 & 0.501 \\
        \rowcolor{LightGray}ConvE~\cite{dettmers2018convolutional} & 504 & 0.942 & 0.955 & 51 & 0.657 & 0.831 \\
        ComplEx~\cite{trouillon2016complex} & -- & 0.941 & 0.947 &  -- & 0.692 & 0.84\\
        \rowcolor{LightGray}RotatE~\cite{sun2019rotate} & 309 & \textbf{0.949} &
        \textbf{0.959}  & \textbf{40} &  \textbf{0.797} &
        \textbf{0.884} \\ 
        MDE  & \textbf{118} & 0.871 & 0.956 & 49 & 0.652 & 0.857\\ 
    \end{tabular}
    }
    \caption{Results on WN18 and FB15k. Best results are in bold.}
    \label{tab:wn18-fb15k}
\end{table}%

\subsection{Link Prediction Results}

Table~\ref{tab:wn18-fb15k} summarizes our results on FB15k and WN18.
It shows that MDE performs almost like RotatE and outperforms other state-of-the-art models in MR and Hit@10 tests. Table~\ref{tab:wn18RR-fb15k-237} shows the results of the experiments on FB15k-237 and WN18RR, these results follow the same pattern as the ones reported in Table~\ref{tab:wn18-fb15k}. 



Due to the existence of hard limits in the limit-based loss, the mean rank in MDE is lower than most of the other methods. It is noticeable that the addition of independent vectors in the model does not decrease the mean rank of the model, whereas in models with high vector dimensions, the MR and MRR results are unbalanced.
For example, for ComplEx and ConvE which both use a vector dimension of 200, the MRR is significant but the MR is high (which is not suitable). On a different note, RotatE mitigates this issue with the application of a high number of negative samples per positive samples.

\begin{table}
\centering
\resizebox{\columnwidth}{!}{
    \begin{tabular}{c|ccc|cccc}
        & &\textbf{WN18RR} & & & \textbf{FB15k-237} & \\
        Model & MR & MRR & Hit@10 & MR & MRR & Hit@10 \\
        \hline
    DistMult~\cite{dismult} & 5110 & 0.43 & 0.49 & 254 & 0.241 & 0.419 \\
    \rowcolor{LightGray}ComplEx~\cite{trouillon2016complex} & 5261 & 0.44 & 0.51 & 339 & 0.247  & 0.428 \\
    ConvE~\cite{dettmers2018convolutional} & 5277 & 0.46 & 0.48 & 246 & 0.316 & 0.491 \\
    \rowcolor{LightGray}RotatE~\cite{sun2019rotate} & 3340 & \textbf{0.476} & \textbf{0.571} & \textbf{177} & 0.338 & \textbf{0.533} \\
    MDE & \textbf{2629} & 0.457 & 0.536 & 189 & 0.288 & 0.484 \\
    \rowcolor{LightGray}MDE$_{NN}$ &  3165 &  0.432 &  0.531 & - & - & -
    \\
    MDE$_{adv}$ & 3219 & 0.458 & 0.560 & 203 & \textbf{0.344} & 0.531
    \\
    \end{tabular}
    }
    \caption{Results on WN18RR and FB15k-237. Best ones are in bold.}
\label{tab:wn18RR-fb15k-237}
\end{table}

The comparison of our model to other state-of-the-art methods in Table~\ref{tab:wn18RR-fb15k-237}, shows the competitive performance of MDE and MDE$_{adv}$.
It is observable that in the MDE tests with only one negative sample per positive sample and using vector sizes between 50 to 200, MDE challenges models with relatively large embedding dimensions (1000) and high number of negative samples (up to 1024). In the ablation study presented in~\cite{sun2019rotate}, we notice that RotatE (with the margin-based ranking criterion, and without self-adversarial negative sampling) produces a Hit@10 score of 0.476 on FB15k-237, which is lower than MDE score.

The adaptation of self-adversarial negative sampling in MDE improves the Hit@10 ranking and the MRR score of the model. This improvement is more significant on the FB15k-237 rather than on the WN18RR, as there is a greater number of relations and entities in FB15k-237 and the self-adversarial negative sampling increases the coverage of different combinations of entities in the training.
We also observe on the FB15-237 benchmark, that MDE$_{adv}$ outperforms previous models on the MRR score since it exists more relations with composition pattern in this dataset than in the WN18RR dataset.







We include each of the terms in MDE as we hypothesize that each one contributes to the generalization power of the model. Practically, we verify this approach in the following section.


\subsection{Ablation Study}
To better understand the role of each term in the score function of MDE, we embark two ablation experiments. First, we train MDE using one of the terms alone, and observe the link prediction performance of each term in the filtered setting. In the second experiment, we remove one of the terms at a time and test the effect of the removal of that term on the model after 100 iterations.

\begin{table}
\centering
\resizebox{\columnwidth}{!}{
    \begin{tabular}{c|ccc|ccc}
        & &\textbf{WN18RR} & & & \textbf{FB15k-237} & \\
        Individual Term & MR & MRR & Hit@10 & MR & MRR & Hit@10 \\
        \hline
    S$_1$ & 3137 & 0.184 & 0.447 & 187 & 0.260 & 0.454 \\
    \rowcolor{LightGray}S$_2$ & 8063 & 0.283 & 0.376 & 439 & 0.204 & 0.342 \\
    S$_3$ & 3153 & 0.183 & 0.449 & \textbf{186} & 0.258 & 0.455 \\
    \rowcolor{LightGray}S$_4$ & \textbf{2245} & \textbf{0.323} & \textbf{0.467} & 220 & \textbf{0.273} & \textbf{0.462} \\
    \end{tabular}
    }
    \caption{Results of each individual term in MDE on WN18RR and FB15k-237. Best results are in bold.}
\label{tab:ab1-wn18RR-fb15k-237}
\end{table}

Table~\ref{tab:ab1-wn18RR-fb15k-237} summarizes the results of the first experiment on WN18RR and FB15k-237. We can see that S$_4$ outperforms the other terms while S$_1$ and S$_3$ perform very similar on these two datasets. Between the four terms, S$_2$ performs the worst since most of the relations in the test datasets follow an antisymmetric pattern and S$_2$ is not efficient in modeling them.

\begin{table}
\centering
\resizebox{\columnwidth}{!}{
    \begin{tabular}{c|ccc|ccc}
        & &\textbf{WN18RR} & & & \textbf{WIN18} & \\
        Removed Term & MR & MRR & Hit@10 & MR & MRR & Hit@10 \\
        \hline
    S$_1$ & 3983 & 0.417 & \textbf{0.501} & \textbf{113} & 0.838 & \textbf{0.946} \\
    \rowcolor{LightGray}S$_2$ & \textbf{3727} & 0.358 & 0.490 & 131 & 0.823 & 0.943 \\
    S$_3$ & 3960 & 0.427 & 0.499 & 161 & 0.850 & 0.943 \\
    \rowcolor{LightGray}S$_4$ & 3921 & 0.366 & 0.478 & 163 & 0.705 & 0.929 \\
    $None$  & 3985 & \textbf{0.428} & \textbf{0.501} & 151 & \textbf{0.844} & \textbf{0.946} \\
    \end{tabular}
    }
    \caption{Results of MDE after 100 iterations when removing one of the terms. Best results are in bold.}
\label{tab:ab2-wn18RR-WN18}
\end{table}

Table~\ref{tab:ab2-wn18RR-WN18} shows the results of the second experiment. The evaluations on WN18RR and WN18 show that the removal of S$_4$ has the most negative effect on the performance of MDE. The removal of S$_1$ that was one of the good performing terms in the last experiment has the least effect. Nevertheless, S$_1$ improves the MRR in the MDE. Also, when we remove S$_2$, the MRR and Hit@10 are negatively influenced, indicating that it exists cases that S$_2$ performs better than the other terms, although, in the individual tests, it performed the worst between all the terms.


\section{Conclusion}

In this study, we created a model based on the generation of several independent vectors for each entity and relation that overrides the expressiveness restrictions of most of the embedding models.
To our knowledge beside MDE and RotatE, other existing KG embedding approaches are unable to allow modeling of all the three relation patterns.
We framed MDE into a Neural Network structure and validated our contributions via both theoretical proofs and empirical results.

We demonstrated that with multiple views to translation embeddings and by using independent vectors (it was previously supposed to cause poor performance~\cite{trouillon2017knowledge,kazemi2018simple}), a model can perform solidly in the link prediction task.
Our experimental results confirm the competitive performances of MDE in MR and Hit@10 on the benchmark datasets. Particularly, MDE outperforms all the current state-of-the-art models for the benchmark of composition relation patterns. 

\clearpage 

\section*{Acknowledgement}
This study is partially supported by the MLwin project (Maschinelles Lernen mit Wissensgraphen, grant 01IS18050F of the Federal Ministry of Education and Research of Germany)\footnote{\url{https://mlwin.de/}}, by Fraunhofer IAIS, and by the ADAPT Centre for Digital Content Technology funded under the SFI Research Centres Programme (Grant 13/RC/2106) and co-funded under the European Regional Development Fund. The authors would also like to thank Michael Galkin for running a part of MDE$_{adv}$ experiments.







\bibliography{references}
\bibliographystyle{ecai}


\newpage
\appendix

\end{document}